\documentclass[11pt,a4paper]{article}
\usepackage[hyperref]{acl2021}
\usepackage{times}
\usepackage{amsmath}
\usepackage{amssymb}
\usepackage{amsthm, bm}
\usepackage{autobreak}
\usepackage{mathrsfs}
\usepackage{latexsym}
\usepackage{booktabs}
\usepackage{graphicx}
\usepackage{subfigure}
\usepackage{footnote}
\usepackage{color}
\usepackage{epstopdf}
\usepackage{hyperref}

\usepackage{CJK}
\usepackage{url}
\usepackage{float}
\usepackage{microtype}
\usepackage{makecell}
\usepackage{multirow}
\usepackage{hyperref}

\aclfinalcopy 
\def\aclpaperid{1704} 

\title{SMedBERT: A Knowledge-Enhanced Pre-trained Language Model with Structured Semantics for Medical Text Mining}
\author{Taolin Zhang$^{1,2,3}$ \thanks{\ \ T. Zhang and Z. Cai contributed equally to this work.}, Zerui Cai$^{4}$\footnotemark[1], Chengyu Wang$^{2}$, Minghui Qiu$^{2}$ \\ \textbf{Bite Yang}$^{5}$, \textbf{Xiaofeng He}$^{4,6}$\thanks{\ \ Corresponding author.}\\
$^1$ School of Software Engineering, East China Normal University
$^2$ Alibaba Group\\
$^3$ Shanghai Key Laboratory of Trsustworthy Computing \\
$^4$ School of Computer Science and Technology, East China Normal University
$^5$ DXY \\
$^6$ Shanghai Research Institute for Intelligent Autonomous Systems\\
 {\tt  zhangtl0519@gmail.com, zrcai\_flow@126.com, yangbt@dxy.cn} \\
 {\tt \{chengyu.wcy, minghui.qmh\}@alibaba-inc.com, hexf@cs.ecnu.edu.cn} \\
 }

\date{}

\begin{document}
\maketitle
\begin{CJK}{UTF8}{gbsn}
\begin{abstract}
Recently, the performance of Pre-trained Language Models (PLMs) has been significantly improved by injecting knowledge facts to enhance their abilities of language understanding. For medical domains, the background knowledge sources are especially useful, due to the massive medical terms and their complicated relations are difficult to understand in text.
In this work, we introduce SMedBERT, a medical PLM trained on large-scale medical corpora, incorporating deep structured semantics knowledge from neighbours of linked-entity.
In SMedBERT, the mention-neighbour hybrid attention is proposed to learn heterogeneous-entity information, which infuses the semantic representations of entity types into the homogeneous neighbouring entity structure.
Apart from knowledge integration as external features, we propose to employ the neighbors of linked-entities in the knowledge graph as additional global contexts of text mentions, allowing them to communicate via shared neighbors, thus enrich their semantic representations.
Experiments demonstrate that SMedBERT significantly outperforms strong baselines in various knowledge-intensive Chinese medical tasks. It also improves the performance of other tasks such as question answering, question matching and natural language inference.\footnote{The code and pre-trained models will be available at \url{https://github.com/MatNLP/SMedBERT}.}
\end{abstract}

\begin{figure}
\centering
\includegraphics[height=4.5cm, width=7.5cm]{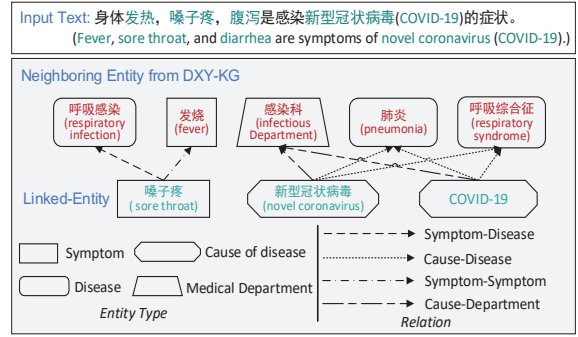}
\caption{Example of neighboring entity information in medical text. (Best viewed in color)}
\label{figure_motivation_example}
\end{figure}

\section{Introduction}
Pre-trained Language Models (PLMs) learn effective context representations with self-supervised tasks, spotlighting in various NLP tasks \citep{DBLP:conf/aaai/WangKMYTACFMMW19,DBLP:conf/acl/NanGSL20, DBLP:conf/acl/LiuGFYCJLD20}.
In addition, Knowledge-Enhanced PLMs (KEPLMs) \citep{DBLP:conf/acl/ZhangHLJSL19,DBLP:conf/aaai/LiuZ0WJD020,DBLP:journals/corr/abs-1911-06136} further benefit language understanding by grounding these PLMs with high-quality, human-curated knowledge facts, which are difficult to learn from raw texts.

In the literatures, a majority of KEPLMs \citep{DBLP:journals/corr/abs-2009-02835, DBLP:conf/aaai/HayashiHXN20, DBLP:conf/coling/SunSQGHHZ20} inject information of entities corresponding to mention-spans from Knowledge Graphs (KGs) into contextual representations.
However, those KEPLMs only utilize linked-entity in the KGs as auxiliary information, which pay little attention to the neighboring structured semantics information of the entity linked with text mentions.
In the medical context, there exist complicated domain knowledge such as relations and medical facts among medical terms \citep{rotmensch2017learning, DBLP:journals/artmed/LiWYWLJSTCWL20}, 
which are difficult to model using previous approaches.
To address this issue, we consider leveraging structured semantics knowledge in medical KGs from the two aspects.
(1) Rich semantic information from neighboring structures of linked-entities, such as entity types and relations, are highly useful for medical text understanding.
As in Figure \ref{figure_motivation_example}, ``新型冠状病毒'' (novel coronavirus) can be the cause of many diseases, such as ``肺炎'' (pneumonia) and ``呼吸综合征'' (respiratory syndrome). \footnote{Although we focus on Chinese medical PLMs here. The proposed method can be easily adapted to other languages, which is beyond the scope of this work.}
(2) Additionally, we leverage neighbors of linked-entity as global ``contexts'' to complement plain-text contexts used in 
\citep{DBLP:journals/corr/abs-1301-3781, DBLP:conf/emnlp/PenningtonSM14}. The structure knowledge contained in neighbouring entities can act as the ``knowledge bridge'' between mention-spans, facilitating the interaction of different mention representations.
Hence, PLMs can learn better representations for rare medical terms.



In this paper, we introduce SMedBERT, a KEPLM pre-trained over large-scale medical corpora and medical KGs.
To the best of our knowledge, SMedBERT is the first PLM with structured semantics knowledge injected in the medical domain.
Specifically, the contributions of SMedBERT mainly include two modules:

\noindent\textbf{Mention-neighbor Hybrid Attention:} We fuse the embeddings of the node and type of linked-entity neighbors into contextual target mention representations. The type-level and node-level attentions help to learn the importance of entity types and the neighbors of linked-entity, respectively, in order to reduce the knowledge noise injected into the model.
The type-level attention transforms the homogeneous node-level attention into a heterogeneous learning process of neighboring entities.

\noindent\textbf{Mention-neighbor Context Modeling:} We propose two novel self-supervised learning tasks for promoting interaction between mention-span and corresponding global context, namely masked neighbor modeling and masked mention modeling. The former enriches the representations of ``context'' neighboring entities based on the well trained ``target word'' mention-span, while the latter focuses on gathering those information back from neighboring entities to the masked target like low-frequency mention-span which is poorly represented \citep{DBLP:conf/acl/TurianRB10}.

In the experiments, we compare SMedBERT against various strong baselines, including mainstream KEPLMs pre-trained over our medical resources.
The underlying medical NLP tasks include: named entity recognition, relation extraction, question answering, question matching and natural language inference.
The results show that SMedBERT consistently outperforms all the baselines on these tasks.

\section{Related Work}
\noindent\textbf{PLMs in the Open Domain.}
PLMs have gained much attention recently, proving successful for boosting the performance of various NLP tasks \citep{DBLP:journals/corr/abs-2003-08271}.
Early works on PLMs focus on feature-based approaches to transform words into distributed representations \citep{DBLP:conf/icml/CollobertW08, DBLP:conf/nips/MikolovSCCD13, DBLP:conf/emnlp/PenningtonSM14, DBLP:conf/naacl/PetersNIGCLZ18}.
BERT \citep{DBLP:conf/naacl/DevlinCLT19} (as well as its robustly optimized version RoBERTa \citep{DBLP:journals/corr/abs-1907-11692}) employs bidirectional transformer encoders \citep{DBLP:conf/nips/VaswaniSPUJGKP17} and self-supervised tasks to generate context-aware token representations.
Further improvement of performances mostly based on the following three types of techniques, including self-supervised tasks \citep{DBLP:journals/tacl/JoshiCLWZL20}, transformer encoder architectures \citep{DBLP:conf/nips/YangDYCSL19} and multi-task learning \citep{DBLP:conf/acl/LiuHCG19}.

\noindent\textbf{Knowledge-Enhanced PLMs.}
\label{KEPLMs_type}
As existing BERT-like models only learn knowledge from plain corpora, various works have investigated how to incorporate knowledge facts to enhance the language understanding abilities of PLMs.
KEPLMs are mainly divided into the following three types.
(1) Knowledge-enhanced by Entity Embedding:
ERNIE-THU \citep{DBLP:conf/acl/ZhangHLJSL19} and KnowBERT \citep{DBLP:conf/emnlp/PetersNLSJSS19} inject linked-entity as heterogeneous features learned by KG embedding algorithms such as TransE \citep{DBLP:conf/nips/BordesUGWY13}.
(2) Knowledge-enhanced by Entity Description: E-BERT \citep{DBLP:journals/corr/abs-2009-02835} and KEPLER \citep{DBLP:journals/corr/abs-1911-06136} add extra description text of entities to enhance semantic representation.
(3) Knowledge-enhanced by Triplet Sentence: K-BERT \citep{DBLP:conf/aaai/LiuZ0WJD020} and CoLAKE \citep{DBLP:conf/coling/SunSQGHHZ20} convert triplets into sentences and insert them into the training corpora without pre-trained embedding.
Previous studies on KG embedding \citep{DBLP:conf/conll/NguyenSQJ16, DBLP:conf/esws/SchlichtkrullKB18} have shown that utilizing the surrounding facts of entity can obtain more informative embedding, which is the focus of our work.

\begin{figure*}
\centering
\includegraphics[height=7cm, width=16cm]{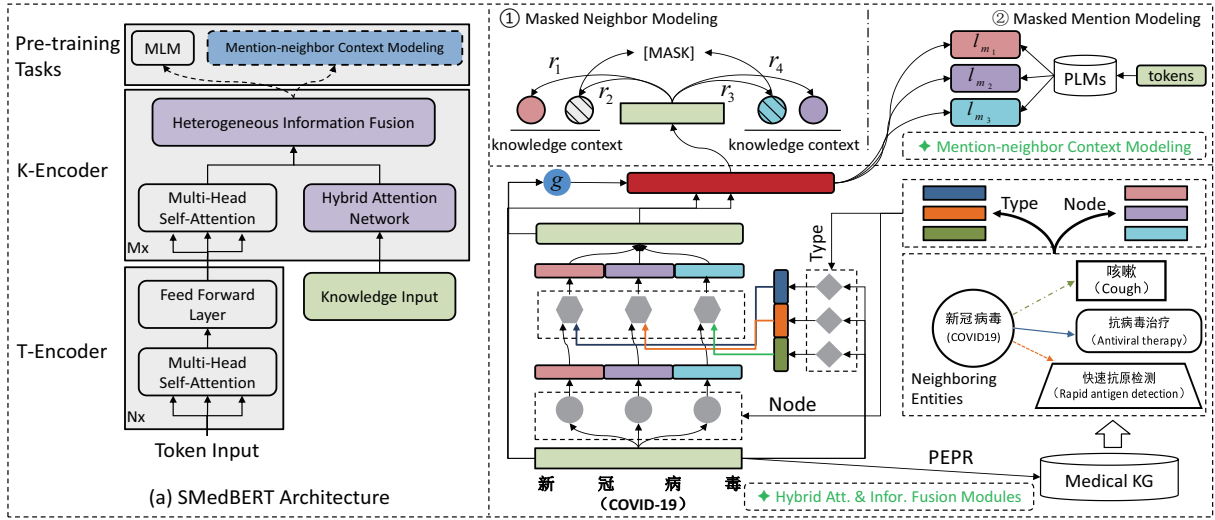}
\caption{Model overview of SMedBERT. The left part is our model architecture and the right part is the details of our model including hybrid attention network and mention-neighbor context modeling pre-training tasks.}
\label{paper_model}
\end{figure*}

\noindent\textbf{PLMs in the Medical Domain.}
PLMs in the medical domain can be generally divided into three categories.
(1) BioBERT \citep{DBLP:journals/bioinformatics/LeeYKKKSK20}, BlueBERT \citep{DBLP:conf/bionlp/PengYL19}, SCIBERT \citep{DBLP:conf/emnlp/BeltagyLC19} and ClinicalBert \citep{DBLP:journals/corr/abs-1904-05342} apply continual learning on medical domain texts, such as PubMed abstracts, PMC full-text articles and MIMIC-III clinical notes.
(2) PubMedBERT \citep{DBLP:journals/corr/abs-2007-15779} learns weights from scratch using PubMed data to obtain an in-domain vocabulary, alleviating the out-of-vocabulary (OOV) problem.
This training paradigm needs the support of large-scale domain data and resources. 
(3) Some other PLMs use domain self-supervised tasks for pre-training. For example, MC-BERT \citep{DBLP:journals/corr/abs-2008-10813} masks Chinese medical entities and phrases to learn complex structures and concepts.
DiseaseBERT \citep{DBLP:conf/emnlp/HeZZCC20} leverages the medical terms and its category as the labels to pre-train the model.
In this paper, we utilize both domain corpora and neighboring entity triplets of mentions to enhance the learning of medical language representations.

\section{The SMedBERT Model}
\subsection{Notations and Model Overview}
In the PLM, we denote the hidden feature of each token $\{w_{1},..., w_{N}\}$ as $\{h_1, h_2,...,h_N\}$ where $N$ is the maximum input sequence length and the total number of pre-training samples as $M$.
Let $E$ be the set of mention-span $e_m$ in the training corpora.
Furthermore, the medical KG consists of the entities set $\mathcal{E}$ and the relations set $\mathcal{R}$.
The triplet set is $S=\{(h, r, t) \mid h \in \mathcal{E}, r \in \mathcal{R}, t \in \mathcal{E}\}$, where $h$ is the head entity with relation $r$ to the tail entity $t$.
The embeddings of entities and relations trained on KG by TransR \citep{DBLP:conf/aaai/LinLSLZ15} are represented as $\Gamma_{ent}$ and $\Gamma_{rel}$, respectively.
The neighboring entity set recalled from KG by $e_m$ is denoted as $\mathcal{N}_{e_m}=\{e_m^{1}, e_m^{2}, ...,e_m^{K}\}$ where $K$ is the threshold of our PEPR algorithm.
We denote the number of entities in the KG as $Z$.
The dimensions of the hidden representation in PLM and the KG embeddings are $d_1$ and $d_2$, respectively.

The main architecture of the our model is shown in Figure \ref{paper_model}. 
SMedBERT mainly includes three components: (1) Top-K entity sorting determine which K neighbour entities to use for each mention. (2) Mention-neighbor hybrid attention aims to infuse the structured semantics knowledge into encoder layers, which includes type attention, node attention and gated position infusion module. (3) Mention-neighbor context modeling includes masked neighbor modeling and masked mention modeling aims to promote mentions to leverage and interact with neighbour entities.




\subsection{Top-K Entity Sorting}
Previous research shows that simple neighboring entity expansion may induce knowledge noises during PLM training \citep{DBLP:conf/aaai/WangKMYTACFMMW19}.
In order to recall the most important neighboring entity set from the KG for each mention, we extend the Personalized PageRank (PPR) \citep{ilprints422} algorithm to filter out trivial entities.~\footnote{We name our algorithm to be Personalized Entity PageRank, abbreviated as PEPR.}
Recall that the iterative process in PPR is
${V}_{i} = (1-\alpha) {A} \cdot {V}_{i-1}+\alpha {P}$
where $A$ is the normalized adjacency matrix, $\alpha$ is the damping factor, $P$ is uniformly distributed jump probability vector, and $V$ is the iterative score vector for each entity.

PEPR specifically focuses on learning the weight for the target mention span in each iteration.
It assigns the span $e_m$ a higher jump probability 1 in $P$ with the remaining as ${\frac{1}{{Z}}}$.
It also uses the entity frequency to initialize the score vector $V$:
\begin{equation}
V_{e_m}=\left\{\begin{array}{ccc}
\frac{t_{e_m}}{T} && e_m \in E \\ [1.5mm]
\frac{1}{M} && e_m \notin E
\end{array}\right.
\end{equation}
where $\mathrm{T}$ is the sum of frequencies of all entities.
$t_{e_{m}}$ is the frequency of $e_m$ in the corpora.
After sorting, we select the top-$K$ entity set $\mathcal{N}_{e_m}$.

\subsection{Mention-neighbor Hybrid Attention}
Besides the embeddings of neighboring entities,
SMedBERT integrates the type information of medical entities to further enhance semantic representations of mention-span.


\subsubsection{Neighboring Entity Type Attention}
Different types of neighboring entities may have different impacts.
Given a specific mention-span $e_m$, we compute the neighboring entity type attention.
Concretely, we calculate hidden representation of each entity type $\tau$ as $h_{\tau}=\sum_{e_m^{i}\in E^{\tau}_m} h_{e_m^{i}}$. $E^{\tau}_m$ are neighboring entities of $e_m$ with the same type $\tau$ and $h_{e_m^{i}}=\Gamma_{ent}\left(e_m^{i}\right) \in \mathbb{R}^{d_2}$.
\begin{equation}
h_{e_{m}}^{\prime}=\mathcal{LN}\left(\sigma\left( f_{sp}\left(h_{i}, \ldots, h_{j}\right) W_{be} \right)\right)
\end{equation}
where $f_{sp}$ is the self-attentive pooling \citep{DBLP:conf/iclr/LinFSYXZB17} to generate the mention-span representation $h_{e_m} \in \mathbb{R}^{d_1}$ and the $\left(h_{i}, h_{i+1}, \ldots, h_{j}\right)$ is the hidden representation of tokens $(w_i, w_{i+1}, \ldots, w_j)$ in mention-span $e_m$ trained by PLMs.
$h_{e_{m}}^{\prime} \in \mathbb{R}^{d_2}$ is obtained by $\sigma(\cdot)$ non-linear activation function GELU \citep{hendrycks2016gaussian} and the learnable projection matrix $W_{be} \in \mathbb{R}^{d_1 \times d_2}$.
$\mathcal{LN}$ is the LayerNorm function \citep{DBLP:journals/corr/BaKH16}.
Then, we calculate the each type attention weight using the type representation $h_{\tau} \in \mathbb{R}^{d_2}$ and the transformed mention-span representation $h_{e_m}^{\prime}$:
\begin{gather}
\alpha_{\tau}^{\prime}=\tanh \left(h_{e_{m}}^{\prime}W_{t} + h_{\tau}W_{t^{\prime}}\right)W_{a}
\end{gather}
where $W_{t} \in \mathbb{R}^{d_2 \times d_2}$, $W_{t^{\prime}} \in \mathbb{R}^{d_2 \times d_2}$ and $W_{a} \in \mathbb{R}^{d_2\times 1}$.
Finally, the neighboring entity type attention weights $\alpha_{\tau}$ are obtained by normalizing the attention score $\alpha_{\tau}^{\prime}$ among all entity types $\mathcal{T}$.

\subsubsection{Neighboring Entity Node Attention}
Apart from entity type information, different neighboring entities also have different influences.
Specifically, we devise the neighboring entity node attention to capture the different semantic influences from neighboring entities to the target mention span  and reduce the effect of noises.
We calculate the entity node attention using the mention-span representation $h_{e_m}^{\prime}$ and neighboring entities representation $h_{e_m^{i}}$ with entity type $\tau$ as:
\begin{gather}
\beta_{e_m e_m^{i}}^{\prime}=\frac{\left(h_{e_m}^{\prime}W_{q}\right)\left( h_{e_{m}^{i}}W_{k}\right)^{T}}{\sqrt{d_2}} \alpha_{\tau}\\
\beta_{e_m e_m^{i}}=\frac{\exp \left(\beta_{e_m e_m^{i}}^{\prime}\right)}{\sum_{e_m^{i} \in \mathcal{N}_{e_m}} \exp \left(\beta_{e_m e_m^{i}}^{\prime}\right)} 
\end{gather}
where $W_{q} \in \mathbb{R}^{d_2 \times d_2}$ and $W_{k} \in \mathbb{R}^{d_2 \times d_2}$ are the attention weight matrices.

The representations of all neighboring entities in $\mathcal{N}_{e_m}$ are aggregated to $\Bar{h}_{e_{m}}^{\prime} \in \mathbb{R}^{d_2}$:
\begin{gather}
\widehat{h}_{e_{m}}^{\prime}=\sum_{e_m^{i} \in \mathcal{N}_{e_m}} \beta_{e_{m} e_m^{i}}\left(h_{e_m^{i}}W_{v} +b_v \right)\\
\Bar{h}_{e_{m}}^{\prime}=\mathcal{LN}\left(\widehat{h}_{e_{m}}^{\prime} +\left( \sigma\left( \widehat{h}_{e_{m}}^{\prime}W_{l1}+b_{l1}\right)W_{l2}\right)\right)
\end{gather}
where $W_{v} \in \mathbb{R}^{d_2 \times d_2}$, $W_{l1} \in \mathbb{R}^{d_2 \times 4d_2}$, $W_{l2} \in \mathbb{R}^{4d_2 \times d_2}$. $b_v\in \mathbb{R}^{d_2}$ and $b_{l1}\in \mathbb{R}^{4d_2}$ are the bias vectors.
$\Bar{h}_{e_{m}}^{\prime}$ is the mention-neighbor representation from hybrid attention module.

\subsubsection{Gated Position Infusion}

Knowledge-injected representations may divert the texts from its original meanings. We further reduce knowledge noises via gated position infusion:
\begin{gather}
h_{e_{mf}}^{\prime} =\sigma\left(\left[\Bar{h}_{e_{m}}^{\prime} \parallel h_{e_{m}}^{\prime}\right]W_{mf} +b_{mf}\right) \\
\widetilde{h}_{e_{mf}}^{\prime} =\mathcal{LN}(h_{e_{mf}}^{\prime}W_{bp}+b_{bp})
\end{gather}
where $W_{mf} \in \mathbb{R}^{2d_2 \times 2d_2}$, $W_{bp} \in \mathbb{R}^{2d_2 \times d_1}$, $b_{mf} \in \mathbb{R}^{2d_2}$, $b_{bp} \in \mathbb{R}^{d_1}$. $h_{e_{mf}}^{\prime} \in \mathbb{R}^{2d_2}$ is the span-level infusion representation.
``$\parallel$'' means concatenation operation. $\widetilde{h}_{e_{mf}}^{\prime} \in \mathbb{R}^{d_1}$ is the final knowledge-injected representation for mention $e_{m}$.
We generate the output token representation $h_{if}$ by \footnote{We find that restricting the knowledge infusion position to tokens is helpful to improve performance.}:
\begin{gather}
g_{i}=\tanh \left(\left(\left[h_{i} \parallel \widetilde{h}_{e_{mf}}^{\prime}\right]\right)W_{ug}+b_{u g}\right)\\
h_{if}=\sigma\left(\left(\left[h_{i} \parallel g_{i} * \widetilde{h}_{e_{mf}}^{\prime}\right]\right)W_{ex}+b_{ex}\right) + h_{i}
\end{gather}
where $W_{ug}$, $W_{ex} \in \mathbb{R}^{2d_1 \times d_1}$. $b_{ug},$ $b_{ex}\in\mathbb{R}^{d_1}$. ``$*$'' means element-wise multiplication.

\subsection{Mention-neighbor Context Modeling}
To fully exploit the structured semantics knowledge in KG, we further introduce two novel self-supervised pre-training tasks, namely Masked Neighbor Modeling (MNeM) and Masked Mention Modeling (MMeM).


\subsubsection{Masked Neighbor Modeling}
Formally, let $r$ be the relation between the mention-span $e_{m}$ and a neighboring entity $e_{m}^{i}$:
\begin{gather}
h_{mf}=\mathcal{LN}\left(\sigma\left( f_{sp} \left(h_{if}, \ldots, h_{jf}\right)W_{sa}\right)\right)
\end{gather}
where $h_{mf}$ is the mention-span hidden features based on the tokens hidden representation $\left(h_{if}, h_{\left(i+1\right)f},\ldots,h_{jf}\right)$.
$h_r=\Gamma_{rel}\left(r\right) \in \mathbb{R}^{d_2}$ is the relation $r$ representation and $W_{sa} \in  \mathbb{R}^{d_1 \times d_2}$ is a learnable projection matrix.
The goal of MNeM is leveraging the structured semantics in surrounding entities while reserving the knowledge of relations between entities.
Considering the object functions of skip-gram with negative sampling (SGNS) \citep{DBLP:journals/corr/abs-1301-3781} and score function of TransR \citep{DBLP:conf/aaai/LinLSLZ15}:
\begin{gather}
\mathcal{L}_{\mathrm{S}}  = \log f_s(w,c) +
                                k\cdot \mathbb{E}_{c_n\sim P_D}[\log f_s(w,-c_n)]\\
f_{\mathrm{tr}}(h,r,t) = \parallel hM_r+r-tM_r\parallel
\end{gather}
where the $w$ in $\mathcal{L}_{\mathrm{S}}$ is the target word of context $c$.
$f_s$ is the compatibility function measuring how well the target word is fitted into the context. 
Inspired by SGNS, following the general energy-based framework \citep{lecun2006tutorial},
we treat mention-spans in corpora as ``target words'', and neighbors of corresponding entities in KG as ``contexts'' to provide additional global contexts.
We employ the Sampled-Softmax \citep{DBLP:conf/acl/JeanCMB15} as the criterion $\mathcal{L}_{\mathrm{MNeM}}$ for the mention-span $e_m$:
\begin{gather}
\sum_{\mathcal{N}_{e_m}}\log \frac{\exp(f_s(\theta))}{\exp(f_s(\theta))+ K\cdot {\mathbb{E}_{e_n\sim Q(e_n)}}[\exp (f_s(\theta^{\prime}))]}
\end{gather}
where $\theta$ denotes the triplet $(e_m, r, e_m^{i})$, $e_m^{i}\in\mathcal{N}_{e_m}$. $\theta^{\prime}$ is the negative triplets $(e_m, r, e_n)$, and
$e_n$ is negative entity sampled with $Q(e_m^{i})$ detailed in Appendix \ref{model_para_traing_details}.
To keep the knowledge of relations between entities, we define the compatibility function as:
\begin{equation}
f_{s}\left(e_m, r, e_m^{i} \right)=\frac{h_{mf} M_{r}+h_r}{|| h_{mf} M_{r}+h_r ||} \cdot \frac{(h_{e_{m}^{i}} M_{r})^T}{|| h_{e_{m}^{i}} M_{r}||}\mu
\end{equation}
where $\mu$ is a scale factor. Assuming the norms of both $ h_{mf} M_{r}+h_r$ and $h_{e_{m}^{i}} M_{r}$ are 1,we have:
\begin{equation}
f_{s}\left(e_m, r, e_m^{i} \right)=\mu \iff f_{tr}(h_{mf},h_r,h_{e_{m}^{i}})=0
\end{equation}
which indicates the proposed $f_s$ is equivalence with $f_{tr}$.
Because $\mid h_{e_{n}} M_{r}\mid$ needs to be calculated for each $e_{n}$, the computation of the score function $f_{s}$ is costly.
Hence, we transform part of the formula $f_{s}$ as follows:
\begin{equation}
\begin{array}{l}
\left(h_{mf} M_{r}+h_r\right) \cdot\left(h_{e_{n}} M_{r}\right)^{T}= \\
{\left[\begin{array}{ll}
h_{mf} & 1
\end{array}\right]\left[\begin{array}{c}
 M_{r} \\
h_r
\end{array}\right]\left[\begin{array}{c}
 M_{r} \\
h_r
\end{array}\right]^{T}\left[\begin{array}{ll}
h_{e_{n}} & 0
\end{array}\right]^{T}}\\=
\left[\begin{array}{ll}
h_{mf} & 1
\end{array}\right] M_{P_{r}} \left[\begin{array}{ll}
h_{e_{n}}& 0
\end{array}\right]^{T}
\end{array}
\end{equation}
In this way, we eliminate computation of transforming each $h_{e_{n}}$.
Finally, to compensate the offset introduced by the negative sampling function $Q(e_m^{i})$ \citep{DBLP:conf/acl/JeanCMB15}, we complement $f_{s}(e_m, r, e_m^{i})$ as:
\begin{equation}
\frac{{\left[\begin{array}{ll}
h_{mf} & 1
\end{array}\right]} M_{P_{r}}}{\parallel{\left[\begin{array}{ll}
h_{mf} & 1
\end{array}\right]} M_{P_{r}}\parallel} \cdot \frac{\left[\begin{array}{ll}
h_{e_{m}^{i}} & 0
\end{array}\right]}{\parallel h_{e_{m}^{i}}\parallel}\mu -\mu\log Q(e_m^{i})
\end{equation}

\subsubsection{Masked Mention Modeling}
In contrast to MNeM, MMeM transfers the semantic information in neighboring entities back to the masked mention $e_m$.
\begin{gather}
\mathcal{Y}_{m}=\mathcal{LN}\left( \sigma \left(f_{sp}\left(h_{ip},\ldots, h_{jp}\right)W_{sa}\right)\right)
\end{gather}
where $\mathcal{Y}_{m}$ is the ground-truth representation of $e_m$ and $h_{ip}=\Gamma_p(w_i) \in \mathbb{R}^{d_2}$.
$\Gamma_p$ is the pre-trained embedding of BERT in our medical corpora.
The mention-span representation obtained by our model is $h_{mf}$.
For a sample $s$, the loss of MMeM $\mathcal{L}_{\mathrm{MMeM}}$ is calculated
via Mean-Squared Error:
\begin{equation}
\mathcal{L}_{\mathrm{MMeM}}= \sum_{m_i}^{\mathcal{M}_{s}} \parallel h_{m_{i}f}-\mathcal{Y}_{m_i}\parallel^{2}
\end{equation}
where $\mathcal{M}_s$ is the set of mentions of sample $s$. 
\subsection{Training Objective}
In SMedBERT, the training objectives mainly consist of three parts, including the self-supervised loss proposed in previous works and  the mention-neighbor context modeling loss proposed in our work.
Our model can be applied to medical text pre-training directly in different languages as long as high-quality medical KGs can be obtained.
The total loss is as follows:
\begin{gather}
\mathcal{L}_{\mathrm{total}}=\mathcal{L}_{\mathrm{EX}}+\lambda_1\mathcal{L}_{\mathrm{MNeM}}+\lambda_2\mathcal{L}_{\mathrm{MMeM}}
\end{gather}
where $\mathcal{L}_{\mathrm{EX}}$ is the sum of sentence-order prediction (SOP) \citep{DBLP:conf/iclr/LanCGGSS20} and masked language modeling.
$\lambda_1$ and $\lambda_2$ are the hyperparameters.

\section{Experiments}
\subsection{Data Source}
\noindent\textbf{Pre-training Data.}
The pre-training corpora after pre-processing contains 5,937,695 text segments with 3,028,224,412 tokens (4.9 GB).
The KGs embedding trained by TransR \citep{DBLP:conf/aaai/LinLSLZ15} on two trusted data sources, including the Symptom-In-Chinese from OpenKG\footnote{\url{http://www.openkg.cn/dataset/symptom-in-chinese}} and DXY-KG \footnote{\url{https://portal.dxy.cn/}} containing 139,572 and 152,508 entities, respectively. 
The number of triplets in the two KGs are 1,007,818 and 3,764,711.
The pre-training corpora and the KGs are further described in Appendix ~\ref{pre_training_data}.

\noindent\textbf{Task Data.}
We use four large-scale datasets in ChineseBLUE \citep{DBLP:journals/corr/abs-2008-10813} to evaluate our model, which are benchmark of Chinese medical NLP tasks.
Additionally, we test models on four datasets from real application scenarios provided by DXY company \footnote{\url{https://auth.dxy.cn/accounts/login}} and CHIP \footnote{\url{http://www.cips-chip.org.cn:8088/home}}, i.e., Named Entity Recognition (DXY-NER), Relation Extraction (DXY-RE, CHIP-RE) and Question Answer (WebMedQA \citep{DBLP:journals/midm/HeFT19}).
For other information of the downstream datasets, we refer readers to Appendix ~\ref{task_data}.

\subsection{Baselines}
In this work, we compare SMedBERT with general PLMs, domain-specific PLMs and KEPLMs with knowledge embedding injected, pre-trained on our Chinese medical corpora:

\noindent\textbf{General PLMs:} We use three Chinese BERT-style models, namely BERT-base \citep{DBLP:conf/naacl/DevlinCLT19}, BERT-wwm \citep{DBLP:journals/corr/abs-1906-08101} and RoBERTa \citep{DBLP:journals/corr/abs-1907-11692}.
All the weights are initialized from \citep{DBLP:conf/emnlp/CuiC000H20}.

\noindent\textbf{Domain-specific PLMs:}
As very few PLMs in the Chinese medical domain are available, we consider the following models. 
MC-BERT \citep{DBLP:journals/corr/abs-2008-10813} is pre-trained over a Chinese medical corpora via masking different granularity tokens.
We also pre-train BERT using our corpora, denoted as BioBERT-zh.


\noindent\textbf{KEPLMs:} We employ two SOTA KEPLMs continually pre-trained on our medical corpora as our baseline models, including ERNIE-THU \citep{DBLP:conf/acl/ZhangHLJSL19} and KnowBERT \citep{DBLP:conf/emnlp/PetersNLSJSS19}.
For a fair comparison, KEPLMs use other additional resources rather than the KG embedding are excluded (See Section \ref{KEPLMs_type}), and all the baseline KEPLMs are injected by the same KG embedding.

The detailed parameter settings and training procedure are in Appendix \ref{model_para_traing_details}.


\begin{table}[t]
\centering
\begin{footnotesize}
\begin{tabular}{c|c|c|c}
\toprule
\textbf{Model} & \textbf{D1} & \textbf{D2} & \textbf{D3} \\ \midrule
SGNS-char-med & 27.21\% & 27.16\% & 21.72\%  \\
SGNS-word-med & 24.64\% & 24.95\% & 20.37\% \\ 
GLOVE-char-med & 27.24\% & 27.12\% & 21.91\% \\ 
GLOVE-word-med & 24.41\% & 23.89\% & 20.56\% \\ \midrule
BERT-open & 29.79\% & 29.41\% & 21.83\% \\ 
BERT-wwm-open & 29.75\% & 29.55\%  & 21.97\%\\ 
RoBERTa-open & 30.84\% & 30.56\% & 21.98\% \\ \midrule
MC-BERT & 30.63\% & 30.34\% & 22.65\% \\ 
BioBERT-zh & 30.84\% & 30.69\% & 22.71\% \\ 
ERNIE-med & 30.97\% & 30.78\%  & 22.99\%\\ 
KnowBERT-med & 30.95\% & 30.77\% & 23.07\%  \\ \midrule
SMedBERT & \textbf{31.81\%}  & \textbf{32.14\%} & \textbf{24.08}\% \\  \bottomrule
\end{tabular}
\end{footnotesize}
\caption{\label{unsupervised_similarity} Results of unsupervised semantic similarity task.
``med'' refers to models continually pre-trained on medical corpora, and ``open'' means open-domain corpora.
``char' and ``word'' refer to the token granularity of input samples.}
\end{table}

\begin{table*}[t]
\centering
\begin{footnotesize}
\begin{tabular}{cccccccccc}
\toprule
  & \multicolumn{5}{c}{Named Entity Recognition} & \multicolumn{4}{c}{Relation Extraction} \\
\cmidrule(r){2-6} \cmidrule(r){7-10}
Model & \multicolumn{2}{c}{cMedQANER} & \multicolumn{2}{c}{DXY-NER} & \multicolumn{1}{c}{Average} & CHIP-RE  & \multicolumn{2}{c}{DXY-RE} & \multicolumn{1}{c}{Average} \\
\cmidrule(r){2-3} \cmidrule(r){4-5} \cmidrule(r){6-6} \cmidrule(r){7-7}\cmidrule(r){8-9} \cmidrule(r){10-10}
 & Dev & Test & Dev & Test &  Test & Test & Dev & Test & Test \\
\midrule
BERT-open & 80.69\% & 83.12\% & 79.12\% & 79.03\%  & 81.08\%  & 85.86\% & 94.18\% & 94.13\% & 90.00\% \\
BERT-wwm-open & 80.52\% & 83.07\% & 79.48\% & 79.29\% & 81.18\%  & 86.01\% & 94.35\% & 94.38\% & 90.20\%\\
RoBERT-open & 80.92\% & 83.29\% & 79.27\% & 79.33\% & 81.31\%  & 86.19\% & 94.64\% & 94.66\% & 90.43\% \\\midrule
 BioBERT-zh & 80.72\% & 83.38\% & 79.52\% & 79.45\% & 81.42\%  & 86.12\% & 94.54\% & 94.64\% & 90.38\% \\ 
MC-BERT & 81.02\% & 83.46\% & 79.79\% & 79.59\% & 81.53\%  & 86.09\% & 94.74\% & 94.73\% & 90.41\% \\
KnowBERT-med &  81.29\% & 83.75\% & 80.86\% & 80.44\% & 82.10\%  & 86.27\% & 95.05\% & 94.97\% & 90.62\% \\
ERNIE-med & 81.22\% & 83.87\% & 80.82\% & 80.87\%  & 82.37\%  & 86.25\% & 94.98\% & 94.91\% & 90.58\%\\ \midrule
SMedBERT & \textbf{82.23\%} & \textbf{84.75\%} & \textbf{83.06\%} & \textbf{82.94\%} & \textbf{83.85\%}  &  \textbf{86.95\%} & \textbf{95.73\%} & \textbf{95.89\%}&\textbf{91.42\%} \\\bottomrule
\end{tabular}
\end{footnotesize}
\caption{Performance of Named Entity Recognition (NER) and Relation Extraction (RE) tasks in terms of F1. The Development data of CHIP-RE is unreleased in public dataset.}
\label{entity_related_result}
\end{table*}

\begin{table*}[]
\centering
\begin{footnotesize}
\begin{tabular}{cccccccccc}
\toprule
 & \multicolumn{5}{c}{Question Answering} & \multicolumn{2}{c}{Question Matching} & \multicolumn{2}{c}{Natural Lang. Infer.} \\
\cmidrule(r){2-6} \cmidrule(r){7-8} \cmidrule(r){9-10}
Model & \multicolumn{2}{c}{cMedQA} & \multicolumn{2}{c}{WebMedQA} &\multicolumn{1}{c}{Average} & \multicolumn{2}{c}{cMedQQ}  & \multicolumn{2}{c}{cMedNLI} \\
\cmidrule(r){2-3} \cmidrule(r){4-5}\cmidrule(r){6-6} \cmidrule(r){7-8} \cmidrule(r){9-10}
 & Dev & Test & Dev & Test & Test & Dev & Test & Dev & Test \\
\midrule
BERT-open & 72.99\% &  73.82\% &  77.20\% & 79.72\% & 76.77\% & 86.74\% & 86.72\% & 95.52\% & 95.66\% \\
BERT-wwm-open & 72.03\% & 72.96\% & 77.06\% & 79.68\% & 76.32\%  & 86.98\% & 86.82\% & 95.53\% & 95.78\% \\
RoBERT-open & 72.22\% & 73.18\% & 77.18\% & 79.57\% & 76.38\%  & 87.24\% & 86.97\% & 95.87\% & 96.11\%\\ \midrule
 BioBERT-zh & 74.32\% & 75.12\% & 78.04\% & 80.45\% & 77.79\%  & 87.30\% & 87.06\% & 95.89\% & 96.04\% \\
MC-BERT & 74.40\% & 74.46\% & 77.85\% & 80.54\% & 77.50\% & 87.17\% & 87.01\% & 95.81\% & 96.06\%\\
KnowBERT-med & 74.38\% & 75.25\% & 78.20\% & 80.67\% & 77.96\% & 87.25\% & 87.14\% & 95.96\% & 96.03\% \\
ERNIE-med & 74.37\% & 75.22\% & 77.93\% & 80.56\% & 77.89\% & 87.34\% & 87.20\% & 96.02\% & 96.25\% \\ \midrule
SMedBERT & \textbf{75.06\%}& \textbf{76.04\%} & \textbf{79.26\%} & \textbf{81.68\%} & \textbf{78.86\%}&  \textbf{88.13\%} & \textbf{88.09\%} & \textbf{96.64\%} & \textbf{96.88\%} \\\bottomrule
\end{tabular}
\end{footnotesize}
\caption{Performance of Question Answering (QA), Question Matching (QM) and Natural Language Inference (NLI) tasks. The metric of the QA task is Acc@1 and those of QM and NLI are F1.}
\label{entity_unrelated_result}
\end{table*}

\subsection{Intrinsic Evaluation}
To evaluate the semantic representation ability of SMedBERT, we design an unsupervised semantic similarity task.
Specifically, we extract all entities pairs with equivalence relations in KGs as positive pairs.
For each positive pair, we use one of the entity as query entity while the other as positive candidate, which is used to sample other entities as negative candidates. We denote this dataset as \textbf{D1}.
Besides, the entities in the same positive pair often have many neighbours in common.
We select positive pairs with large proportions of common neighbours as \textbf{D2}. 
Additionally, to verify the ability of SMedBERT of enhancing the low-frequency mention representation, we extract all positive pairs that with at least one low-frequency mention as \textbf{D3}.
There are totally 359,358, 272,320 and 41,583 samples for \textbf{D1}, \textbf{D2}, \textbf{D3} respectively.
We describe the details of collecting data and embedding words in Appendix \ref{unsupervised_semantic_similarity}.
In this experiments, we compare SMedBERT with three types of models: classical word embedding methods (\textbf{SGNS} \citep{DBLP:journals/corr/abs-1301-3781}, \textbf{GLOVE} \citep{DBLP:conf/emnlp/PenningtonSM14}), PLMs and KEPLMs.
We compute the similarity between the representation of query entities and all the other entities, retrieving the most similar one. The evaluation metric is top-1 accuracy (Acc@1).

Experiment results are shown in Table \ref{unsupervised_similarity}.
From the results, we observe that:
(1) SMedBERT greatly outperforms all baselines especially on the dataset \textbf{D2 (+1.36\%)}, where most positive pairs have many shared neighbours, demonstrating that ability of SMedBERT to utilize semantic information from the global context.
(2) In dataset \textbf{D3}, SMedBERT improve the performance significantly \textbf{(+1.01\%)}, indicating our model is effective to enhance the representation of low-frequency mentions.

\subsection{Results of Downstream Tasks}
We first evaluate our model in NER and RE tasks that are closely related to entities in the input texts.
Table \ref{entity_related_result} shows the performances on medical NER and RE tasks.
In NER and RE tasks, we can observe from the results: (1) Compared with PLMs trained in open-domain corpora,  KEPLMs with medical corpora and knowledge facts achieve better results. (2) The performance of SMedBERT is greatly improved compared with the strongest baseline in two NER datasets \textbf{(+0.88\%, +2.07\%)}, and \textbf{(+0.68\%, +0.92\%)} on RE tasks.
We also evaluate SMedBERT on QA, QM and NLI tasks and the performance is shown in Table \ref{entity_unrelated_result}.
We can observe that SMedBERT improve the performance consistently on these datasets \textbf{(+0.90\% on QA, +0.89\% on QM and +0.63\% on NLI)}.
In general, it can be seen from Table \ref{entity_related_result} and Table \ref{entity_unrelated_result} that injecting the domain knowledge especially the structured semantics knowledge can improve the result greatly.



\begin{figure}
\flushleft
\includegraphics[height=3.8cm, width=7.8cm]{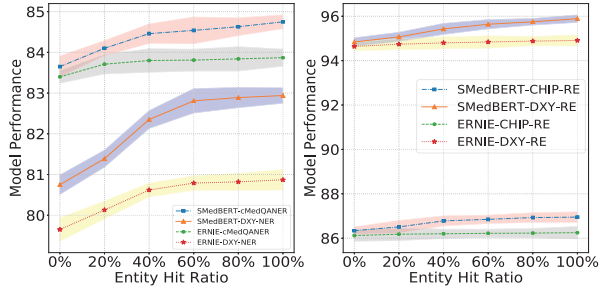}
\caption{Entity hit ratio results of SMedBERT and ERNIE in NER and RE tasks.}
\label{entity_hit_ratio}
\end{figure}

\subsection{Influence of Entity Hit Ratio}
In this experiment, we explore the model performance in NER and RE tasks with different entity hit ratios, which control the proportions of knowledge-enhanced mention-spans in the samples. The average number of mention-spans in samples is about 40.
Figure \ref{entity_hit_ratio} illustrates the performance of SMedBERT and ERNIE-med \citep{DBLP:conf/acl/ZhangHLJSL19}.
From the result, we can observe that: (1) The performance improves significantly at the beginning and then keeps stable as the hit ratio increases, proving the heterogeneous knowledge is beneficial to improve the ability of language understanding and indicating too much knowledge facts are unhelpful to further improve model performance due to the knowledge noise \citep{DBLP:conf/aaai/LiuZ0WJD020}.
(2) Compared with previous approaches, our SMedBERT model improves performance greatly and more stable.

\begin{figure}
\centering
\includegraphics[height=3.5cm,width=6.9cm]{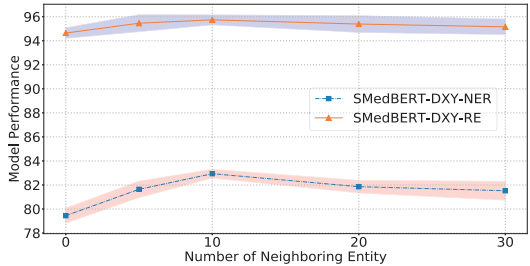}
\caption{The influence of different K values in results.}
\label{K_value_result}
\end{figure}

\subsection{Influence of Neighboring Entity Number}
We further evaluate the model performance under different $K$ over the test set of DXY-NER and DXY-RE.
Figure \ref{K_value_result} shows the the model result with $K=\{5, 10, 20, 30\}$.
In our settings, the SMedBERT can achieve the best performance in different tasks around $K=10$.
The results of SMedBERT show that the model performance increasing first and then decreasing with the increasing of $K$.
This phenomenon also indicates the knowledge noise problem that injecting too much knowledge of neighboring entities may hurt the performance.

\begin{table}
\footnotesize
\flushleft
\begin{tabular}{ccccc}
\toprule
Model & D5 & D6 & D7 & D8 \\\midrule
SMedBERT & \bf 84.75\% & \bf 82.94\% & \bf 86.95\% & \bf 95.89\% \\ 
ERNIE-med & 83.87\% &  80.87\% &  86.25\% &  94.91\% \\  \midrule
\textbf{-} Type Att. & 84.25\% & 81.99\% & 86.61\% & 95.29\% \\
\textbf{-} Hybrid Att.  & 83.71\% & 80.85\% & 86.46\% & 95.20\%\\
\textbf{-} Know. Loss & 84.31\% & 82.12\% & 86.50\% & 95.43\% \\ \bottomrule
\end{tabular}
\caption{Ablation study of SMedBERT on four datasets (testing set). Due to the space limitation, we use the abbreviations ``D5'', ``D6'', ``D7'', and ``D8'' to represent the cMedQANER, DXY-NER, CHIP-RE, and DXY-RE datasets respectively.}
\label{ablation_study}
\end{table}
\subsection{Ablation Study}
In Table \ref{ablation_study}, we choose three important model components for our ablation study and report the test set performance on four datasets of NER and RE tasks that are closely related to entities.
Specifically, the three model components are neighboring entity type attention, the whole hybrid attention module, and mention-neighbor context modeling respectively, which includes two masked language model loss $\mathcal{L}_{\mathrm{MNeM}}$ and $\mathcal{L}_{\mathrm{MMeM}}$.

From the result, we can observe that: (1) Without any of the three mechanisms, our model performance can also perform competitively with the strong baseline ERNIE-med \citep{DBLP:conf/acl/ZhangHLJSL19}.
(2) Note that after removing the hybrid attention module, the performance of our model has the greatest decline, which indicates that injecting rich heterogeneous knowledge of neighboring entities is effective.

\section{Conclusion}
In this work, we address medical text mining tasks with the structured semantics KEPLM proposed named SMedBERT.
Accordingly, we inject entity type semantic information of neighboring entities into node attention mechanism via heterogeneous feature learning process.
Moreover, we treat the neighboring entity structures as additional global contexts to predict the masked candidate entities based on 
mention-spans and vice versa.
The experimental results show the significant improvement of our model on various medical NLP tasks and the intrinsic evaluation.
There are two research directions that can be further explored: (1) Injecting deeper knowledge by using ``farther neighboring'' entities as contexts;
(2) Further enhancing Chinese medical long-tail entity semantic representation.

\section*{Acknowledgements}
We would like to thank anonymous reviewers for their valuable comments.
This work is supported by the National Key Research and Development Program of China under Grant No. 2016YFB1000904, and Alibaba Group through Alibaba Research Intern Program.
\bibliographystyle{acl_natbib}
\bibliography{anthology,acl2021}

\begin{thebibliography}{50}
\expandafter\ifx\csname natexlab\endcsname\relax\def\natexlab#1{#1}\fi

\bibitem[{Ba et~al.(2016)Ba, Kiros, and Hinton}]{DBLP:journals/corr/BaKH16}
Lei~Jimmy Ba, Jamie~Ryan Kiros, and Geoffrey~E. Hinton. 2016.
\newblock \href {http://arxiv.org/abs/1607.06450} {Layer normalization}.
\newblock \emph{CoRR}, abs/1607.06450.

\bibitem[{Beltagy et~al.(2019)Beltagy, Lo, and
  Cohan}]{DBLP:conf/emnlp/BeltagyLC19}
Iz~Beltagy, Kyle Lo, and Arman Cohan. 2019.
\newblock \href {https://doi.org/10.18653/v1/D19-1371} {Scibert: {A} pretrained
  language model for scientific text}.
\newblock In \emph{EMNLP}, pages 3613--3618.

\bibitem[{Bordes et~al.(2013)Bordes, Usunier, Garc{\'{\i}}a{-}Dur{\'{a}}n,
  Weston, and Yakhnenko}]{DBLP:conf/nips/BordesUGWY13}
Antoine Bordes, Nicolas Usunier, Alberto Garc{\'{\i}}a{-}Dur{\'{a}}n, Jason
  Weston, and Oksana Yakhnenko. 2013.
\newblock \href
  {http://papers.nips.cc/paper/5071-translating-embeddings-for-modeling-multi-relational-data}
  {Translating embeddings for modeling multi-relational data}.
\newblock In \emph{NIPS}, pages 2787--2795.

\bibitem[{Collobert and Weston(2008)}]{DBLP:conf/icml/CollobertW08}
Ronan Collobert and Jason Weston. 2008.
\newblock \href {https://doi.org/10.1145/1390156.1390177} {A unified
  architecture for natural language processing: deep neural networks with
  multitask learning}.
\newblock In \emph{ICML}, pages 160--167.

\bibitem[{Cui et~al.(2020)Cui, Che, Liu, Qin, Wang, and
  Hu}]{DBLP:conf/emnlp/CuiC000H20}
Yiming Cui, Wanxiang Che, Ting Liu, Bing Qin, Shijin Wang, and Guoping Hu.
  2020.
\newblock \href {https://doi.org/10.18653/v1/2020.findings-emnlp.58}
  {Revisiting pre-trained models for chinese natural language processing}.
\newblock In \emph{EMNLP}, pages 657--668.

\bibitem[{Cui et~al.(2019)Cui, Che, Liu, Qin, Yang, Wang, and
  Hu}]{DBLP:journals/corr/abs-1906-08101}
Yiming Cui, Wanxiang Che, Ting Liu, Bing Qin, Ziqing Yang, Shijin Wang, and
  Guoping Hu. 2019.
\newblock \href {https://arxiv.org/abs/1906.08101} {Pre-training with whole
  word masking for chinese {BERT}}.
\newblock \emph{CoRR}, abs/1906.08101.

\bibitem[{Devlin et~al.(2019)Devlin, Chang, Lee, and
  Toutanova}]{DBLP:conf/naacl/DevlinCLT19}
Jacob Devlin, Ming{-}Wei Chang, Kenton Lee, and Kristina Toutanova. 2019.
\newblock \href {https://doi.org/10.18653/v1/n19-1423} {{BERT:} pre-training of
  deep bidirectional transformers for language understanding}.
\newblock In \emph{NAACL}, pages 4171--4186.

\bibitem[{Gu et~al.(2020)Gu, Tinn, Cheng, Lucas, Usuyama, Liu, Naumann, Gao,
  and Poon}]{DBLP:journals/corr/abs-2007-15779}
Yu~Gu, Robert Tinn, Hao Cheng, Michael Lucas, Naoto Usuyama, Xiaodong Liu,
  Tristan Naumann, Jianfeng Gao, and Hoifung Poon. 2020.
\newblock \href {http://arxiv.org/abs/2007.15779} {Domain-specific language
  model pretraining for biomedical natural language processing}.
\newblock \emph{CoRR}, abs/2007.15779.

\bibitem[{Hayashi et~al.(2020)Hayashi, Hu, Xiong, and
  Neubig}]{DBLP:conf/aaai/HayashiHXN20}
Hiroaki Hayashi, Zecong Hu, Chenyan Xiong, and Graham Neubig. 2020.
\newblock \href {https://aaai.org/ojs/index.php/AAAI/article/view/6298} {Latent
  relation language models}.
\newblock In \emph{AAAI}, pages 7911--7918.

\bibitem[{He et~al.(2019)He, Fu, and Tu}]{DBLP:journals/midm/HeFT19}
Junqing He, Mingming Fu, and Manshu Tu. 2019.
\newblock \href {https://doi.org/10.1186/s12911-019-0761-8} {Applying deep
  matching networks to chinese medical question answering: a study and a
  dataset}.
\newblock \emph{{BMC} Medical Informatics Decis. Mak.}, 19-S(2):91--100.

\bibitem[{He et~al.(2020)He, Zhu, Zhang, Chen, and
  Caverlee}]{DBLP:conf/emnlp/HeZZCC20}
Yun He, Ziwei Zhu, Yin Zhang, Qin Chen, and James Caverlee. 2020.
\newblock \href {https://doi.org/10.18653/v1/2020.emnlp-main.372} {Infusing
  disease knowledge into {BERT} for health question answering, medical
  inference and disease name recognition}.
\newblock In \emph{EMNLP}, pages 4604--4614.

\bibitem[{Hendrycks and Gimpel(2016)}]{hendrycks2016gaussian}
Dan Hendrycks and Kevin Gimpel. 2016.
\newblock \href {https://arxiv.org/pdf/1606.08415.pdf} {Gaussian error linear
  units (gelus)}.
\newblock \emph{arXiv:1606.08415}.

\bibitem[{Huang et~al.(2019)Huang, Altosaar, and
  Ranganath}]{DBLP:journals/corr/abs-1904-05342}
Kexin Huang, Jaan Altosaar, and Rajesh Ranganath. 2019.
\newblock \href {http://arxiv.org/abs/1904.05342} {Clinicalbert: Modeling
  clinical notes and predicting hospital readmission}.
\newblock \emph{CoRR}, abs/1904.05342.

\bibitem[{Jaccard(1912)}]{2010THE}
Paul Jaccard. 1912.
\newblock \href
  {https://nph.onlinelibrary.wiley.com/doi/abs/10.1111/j.1469-8137.1912.tb05611.x}
  {The distribution of the flora in the alpine zone.}
\newblock \emph{New Phydvtologist}, 11(2):37--50.

\bibitem[{Jean et~al.(2015)Jean, Cho, Memisevic, and
  Bengio}]{DBLP:conf/acl/JeanCMB15}
S{\'{e}}bastien Jean, KyungHyun Cho, Roland Memisevic, and Yoshua Bengio. 2015.
\newblock \href {https://doi.org/10.3115/v1/p15-1001} {On using very large
  target vocabulary for neural machine translation}.
\newblock In \emph{ACL}, pages 1--10.

\bibitem[{Joshi et~al.(2020)Joshi, Chen, Liu, Weld, Zettlemoyer, and
  Levy}]{DBLP:journals/tacl/JoshiCLWZL20}
Mandar Joshi, Danqi Chen, Yinhan Liu, Daniel~S. Weld, Luke Zettlemoyer, and
  Omer Levy. 2020.
\newblock \href {https://transacl.org/ojs/index.php/tacl/article/view/1853}
  {Spanbert: Improving pre-training by representing and predicting spans}.
\newblock \emph{Trans. Assoc. Comput. Linguistics}, 8:64--77.

\bibitem[{Lan et~al.(2020)Lan, Chen, Goodman, Gimpel, Sharma, and
  Soricut}]{DBLP:conf/iclr/LanCGGSS20}
Zhenzhong Lan, Mingda Chen, Sebastian Goodman, Kevin Gimpel, Piyush Sharma, and
  Radu Soricut. 2020.
\newblock \href {https://openreview.net/forum?id=H1eA7AEtvS} {{ALBERT:} {A}
  lite {BERT} for self-supervised learning of language representations}.
\newblock In \emph{ICLR}.

\bibitem[{LeCun et~al.(2006)LeCun, Chopra, Hadsell, Ranzato, and
  Huang}]{lecun2006tutorial}
Yann LeCun, Sumit Chopra, Raia Hadsell, M~Ranzato, and F~Huang. 2006.
\newblock \href {http://yann.lecun.com/exdb/publis/pdf/lecun-06.pdf} {A
  tutorial on energy-based learning}.
\newblock \emph{Predicting structured data}, 1(0).

\bibitem[{Lee et~al.(2020)Lee, Yoon, Kim, Kim, Kim, So, and
  Kang}]{DBLP:journals/bioinformatics/LeeYKKKSK20}
Jinhyuk Lee, Wonjin Yoon, Sungdong Kim, Donghyeon Kim, Sunkyu Kim, Chan~Ho So,
  and Jaewoo Kang. 2020.
\newblock \href {https://doi.org/10.1093/bioinformatics/btz682} {Biobert: a
  pre-trained biomedical language representation model for biomedical text
  mining}.
\newblock \emph{Bioinform.}, 36(4):1234--1240.

\bibitem[{Li et~al.(2020)Li, Wang, Yan, Wang, Li, Jiang, Sun, Tang, Chang,
  Wang, and Liu}]{DBLP:journals/artmed/LiWYWLJSTCWL20}
Linfeng Li, Peng Wang, Jun Yan, Yao Wang, Simin Li, Jinpeng Jiang, Zhe Sun,
  Buzhou Tang, Tsung{-}Hui Chang, Shenghui Wang, and Yuting Liu. 2020.
\newblock \href {https://doi.org/10.1016/j.artmed.2020.101817} {Real-world data
  medical knowledge graph: construction and applications}.
\newblock \emph{Artif. Intell. Medicine}, 103:101817.

\bibitem[{Lin et~al.(2015)Lin, Liu, Sun, Liu, and
  Zhu}]{DBLP:conf/aaai/LinLSLZ15}
Yankai Lin, Zhiyuan Liu, Maosong Sun, Yang Liu, and Xuan Zhu. 2015.
\newblock \href {http://www.aaai.org/ocs/index.php/AAAI/AAAI15/paper/view/9571}
  {Learning entity and relation embeddings for knowledge graph completion}.
\newblock In \emph{AAAI}, pages 2181--2187.

\bibitem[{Lin et~al.(2017)Lin, Feng, dos Santos, Yu, Xiang, Zhou, and
  Bengio}]{DBLP:conf/iclr/LinFSYXZB17}
Zhouhan Lin, Minwei Feng, C{\'{\i}}cero~Nogueira dos Santos, Mo~Yu, Bing Xiang,
  Bowen Zhou, and Yoshua Bengio. 2017.
\newblock \href {https://openreview.net/forum?id=BJC\_jUqxe} {A structured
  self-attentive sentence embedding}.
\newblock In \emph{ICLR}.

\bibitem[{Liu et~al.(2020{\natexlab{a}})Liu, Gong, Fu, Yan, Chen, Jiang, Lv,
  and Duan}]{DBLP:conf/acl/LiuGFYCJLD20}
Dayiheng Liu, Yeyun Gong, Jie Fu, Yu~Yan, Jiusheng Chen, Daxin Jiang, Jiancheng
  Lv, and Nan Duan. 2020{\natexlab{a}}.
\newblock \href {https://www.aclweb.org/anthology/2020.acl-main.604/} {Rikinet:
  Reading wikipedia pages for natural question answering}.
\newblock In \emph{ACL}, pages 6762--6771.

\bibitem[{Liu et~al.(2020{\natexlab{b}})Liu, Zhou, Zhao, Wang, Ju, Deng, and
  Wang}]{DBLP:conf/aaai/LiuZ0WJD020}
Weijie Liu, Peng Zhou, Zhe Zhao, Zhiruo Wang, Qi~Ju, Haotang Deng, and Ping
  Wang. 2020{\natexlab{b}}.
\newblock \href {https://aaai.org/ojs/index.php/AAAI/article/view/5681}
  {{K-BERT:} enabling language representation with knowledge graph}.
\newblock In \emph{AAAI}, pages 2901--2908.

\bibitem[{Liu et~al.(2019{\natexlab{a}})Liu, He, Chen, and
  Gao}]{DBLP:conf/acl/LiuHCG19}
Xiaodong Liu, Pengcheng He, Weizhu Chen, and Jianfeng Gao. 2019{\natexlab{a}}.
\newblock \href {https://doi.org/10.18653/v1/p19-1441} {Multi-task deep neural
  networks for natural language understanding}.
\newblock In \emph{ACL}, pages 4487--4496.

\bibitem[{Liu et~al.(2019{\natexlab{b}})Liu, Ott, Goyal, Du, Joshi, Chen, Levy,
  Lewis, Zettlemoyer, and Stoyanov}]{DBLP:journals/corr/abs-1907-11692}
Yinhan Liu, Myle Ott, Naman Goyal, Jingfei Du, Mandar Joshi, Danqi Chen, Omer
  Levy, Mike Lewis, Luke Zettlemoyer, and Veselin Stoyanov. 2019{\natexlab{b}}.
\newblock \href {http://arxiv.org/abs/1907.11692} {Roberta: {A} robustly
  optimized {BERT} pretraining approach}.
\newblock \emph{CoRR}, abs/1907.11692.

\bibitem[{Mikolov et~al.(2013{\natexlab{a}})Mikolov, Chen, Corrado, and
  Dean}]{DBLP:journals/corr/abs-1301-3781}
Tom{\'{a}}s Mikolov, Kai Chen, Greg Corrado, and Jeffrey Dean.
  2013{\natexlab{a}}.
\newblock \href {http://arxiv.org/abs/1301.3781} {Efficient estimation of word
  representations in vector space}.
\newblock In \emph{ICLR}.

\bibitem[{Mikolov et~al.(2013{\natexlab{b}})Mikolov, Sutskever, Chen, Corrado,
  and Dean}]{DBLP:conf/nips/MikolovSCCD13}
Tom{\'{a}}s Mikolov, Ilya Sutskever, Kai Chen, Gregory~S. Corrado, and Jeffrey
  Dean. 2013{\natexlab{b}}.
\newblock \href
  {http://papers.nips.cc/paper/5021-distributed-representations-of-words-and-phrases-and-their-compositionality}
  {Distributed representations of words and phrases and their
  compositionality}.
\newblock In \emph{NIPS}, pages 3111--3119.

\bibitem[{Nan et~al.(2020)Nan, Guo, Sekulic, and Lu}]{DBLP:conf/acl/NanGSL20}
Guoshun Nan, Zhijiang Guo, Ivan Sekulic, and Wei Lu. 2020.
\newblock \href {https://www.aclweb.org/anthology/2020.acl-main.141/}
  {Reasoning with latent structure refinement for document-level relation
  extraction}.
\newblock In \emph{ACL}, pages 1546--1557.

\bibitem[{Nguyen et~al.(2016)Nguyen, Sirts, Qu, and
  Johnson}]{DBLP:conf/conll/NguyenSQJ16}
Dat~Quoc Nguyen, Kairit Sirts, Lizhen Qu, and Mark Johnson. 2016.
\newblock \href {https://doi.org/10.18653/v1/k16-1005} {Neighborhood mixture
  model for knowledge base completion}.
\newblock In \emph{CoNLL}, pages 40--50.

\bibitem[{Page et~al.(1999)Page, Brin, Motwani, and Winograd}]{ilprints422}
Lawrence Page, Sergey Brin, Rajeev Motwani, and Terry Winograd. 1999.
\newblock \href {http://ilpubs.stanford.edu:8090/422/} {The pagerank citation
  ranking: Bringing order to the web.}
\newblock Technical Report 1999-66, Stanford InfoLab.

\bibitem[{Peng et~al.(2019)Peng, Yan, and Lu}]{DBLP:conf/bionlp/PengYL19}
Yifan Peng, Shankai Yan, and Zhiyong Lu. 2019.
\newblock \href {https://doi.org/10.18653/v1/w19-5006} {Transfer learning in
  biomedical natural language processing: An evaluation of {BERT} and elmo on
  ten benchmarking datasets}.
\newblock In \emph{BioNLP}, pages 58--65.

\bibitem[{Pennington et~al.(2014)Pennington, Socher, and
  Manning}]{DBLP:conf/emnlp/PenningtonSM14}
Jeffrey Pennington, Richard Socher, and Christopher~D. Manning. 2014.
\newblock \href {https://doi.org/10.3115/v1/d14-1162} {Glove: Global vectors
  for word representation}.
\newblock In \emph{EMNLP}, pages 1532--1543.

\bibitem[{Peters et~al.(2019)Peters, Neumann, IV, Schwartz, Joshi, Singh, and
  Smith}]{DBLP:conf/emnlp/PetersNLSJSS19}
Matthew~E. Peters, Mark Neumann, Robert L.~Logan IV, Roy Schwartz, Vidur Joshi,
  Sameer Singh, and Noah~A. Smith. 2019.
\newblock \href {https://doi.org/10.18653/v1/D19-1005} {Knowledge enhanced
  contextual word representations}.
\newblock In \emph{EMNLP}, pages 43--54.

\bibitem[{Peters et~al.(2018)Peters, Neumann, Iyyer, Gardner, Clark, Lee, and
  Zettlemoyer}]{DBLP:conf/naacl/PetersNIGCLZ18}
Matthew~E. Peters, Mark Neumann, Mohit Iyyer, Matt Gardner, Christopher Clark,
  Kenton Lee, and Luke Zettlemoyer. 2018.
\newblock \href {https://doi.org/10.18653/v1/n18-1202} {Deep contextualized
  word representations}.
\newblock In \emph{NAACL}, pages 2227--2237.

\bibitem[{Qiu et~al.(2020)Qiu, Sun, Xu, Shao, Dai, and
  Huang}]{DBLP:journals/corr/abs-2003-08271}
Xipeng Qiu, Tianxiang Sun, Yige Xu, Yunfan Shao, Ning Dai, and Xuanjing Huang.
  2020.
\newblock \href {http://arxiv.org/abs/2003.08271} {Pre-trained models for
  natural language processing: {A} survey}.
\newblock \emph{CoRR}, abs/2003.08271.

\bibitem[{Rotmensch et~al.(2017)Rotmensch, Halpern, Tlimat, Horng, and
  Sontag}]{rotmensch2017learning}
Maya Rotmensch, Yoni Halpern, Abdulhakim Tlimat, Steven Horng, and David
  Sontag. 2017.
\newblock \href {https://www.nature.com/articles/s41598-017-05778-z} {Learning
  a health knowledge graph from electronic medical records}.
\newblock \emph{Scientific reports}, 7(1):1--11.

\bibitem[{Schlichtkrull et~al.(2018)Schlichtkrull, Kipf, Bloem, van~den Berg,
  Titov, and Welling}]{DBLP:conf/esws/SchlichtkrullKB18}
Michael~Sejr Schlichtkrull, Thomas~N. Kipf, Peter Bloem, Rianne van~den Berg,
  Ivan Titov, and Max Welling. 2018.
\newblock \href {https://doi.org/10.1007/978-3-319-93417-4\_38} {Modeling
  relational data with graph convolutional networks}.
\newblock In \emph{ESWC}, pages 593--607.

\bibitem[{Sun et~al.(2020)Sun, Shao, Qiu, Guo, Hu, Huang, and
  Zhang}]{DBLP:conf/coling/SunSQGHHZ20}
Tianxiang Sun, Yunfan Shao, Xipeng Qiu, Qipeng Guo, Yaru Hu, Xuanjing Huang,
  and Zheng Zhang. 2020.
\newblock \href {https://www.aclweb.org/anthology/2020.coling-main.327/}
  {Colake: Contextualized language and knowledge embedding}.
\newblock In \emph{COLING}, pages 3660--3670.

\bibitem[{Turian et~al.(2010)Turian, Ratinov, and
  Bengio}]{DBLP:conf/acl/TurianRB10}
Joseph~P. Turian, Lev{-}Arie Ratinov, and Yoshua Bengio. 2010.
\newblock \href {https://www.aclweb.org/anthology/P10-1040/} {Word
  representations: {A} simple and general method for semi-supervised learning}.
\newblock In \emph{ACL}, pages 384--394.

\bibitem[{Vaswani et~al.(2017)Vaswani, Shazeer, Parmar, Uszkoreit, Jones,
  Gomez, Kaiser, and Polosukhin}]{DBLP:conf/nips/VaswaniSPUJGKP17}
Ashish Vaswani, Noam Shazeer, Niki Parmar, Jakob Uszkoreit, Llion Jones,
  Aidan~N. Gomez, Lukasz Kaiser, and Illia Polosukhin. 2017.
\newblock \href {http://papers.nips.cc/paper/7181-attention-is-all-you-need}
  {Attention is all you need}.
\newblock In \emph{NIPS}, pages 5998--6008.

\bibitem[{Wang et~al.(2019{\natexlab{a}})Wang, Kapanipathi, Musa, Yu,
  Talamadupula, Abdelaziz, Chang, Fokoue, Makni, Mattei, and
  Witbrock}]{DBLP:conf/aaai/WangKMYTACFMMW19}
Xiaoyan Wang, Pavan Kapanipathi, Ryan Musa, Mo~Yu, Kartik Talamadupula, Ibrahim
  Abdelaziz, Maria Chang, Achille Fokoue, Bassem Makni, Nicholas Mattei, and
  Michael Witbrock. 2019{\natexlab{a}}.
\newblock \href {https://doi.org/10.1609/aaai.v33i01.33017208} {Improving
  natural language inference using external knowledge in the science questions
  domain}.
\newblock In \emph{AAAI}, pages 7208--7215.

\bibitem[{Wang et~al.(2019{\natexlab{b}})Wang, Gao, Zhu, Liu, Li, and
  Tang}]{DBLP:journals/corr/abs-1911-06136}
Xiaozhi Wang, Tianyu Gao, Zhaocheng Zhu, Zhiyuan Liu, Juanzi Li, and Jian Tang.
  2019{\natexlab{b}}.
\newblock \href {http://arxiv.org/abs/1911.06136} {{KEPLER:} {A} unified model
  for knowledge embedding and pre-trained language representation}.
\newblock \emph{CoRR}, abs/1911.06136.

\bibitem[{Winkler(1990)}]{winkler1990string}
William~E Winkler. 1990.
\newblock \href {https://files.eric.ed.gov/fulltext/ED325505.pdf} {String
  comparator metrics and enhanced decision rules in the fellegi-sunter model of
  record linkage.}

\bibitem[{Xu et~al.(2020)Xu, Zhang, and
  Dong}]{DBLP:journals/corr/abs-2003-01355}
Liang Xu, Xuanwei Zhang, and Qianqian Dong. 2020.
\newblock \href {http://arxiv.org/abs/2003.01355} {Cluecorpus2020: {A}
  large-scale chinese corpus for pre-training language model}.
\newblock \emph{CoRR}, abs/2003.01355.

\bibitem[{Yang et~al.(2019)Yang, Dai, Yang, Carbonell, Salakhutdinov, and
  Le}]{DBLP:conf/nips/YangDYCSL19}
Zhilin Yang, Zihang Dai, Yiming Yang, Jaime~G. Carbonell, Ruslan Salakhutdinov,
  and Quoc~V. Le. 2019.
\newblock \href
  {http://papers.nips.cc/paper/8812-xlnet-generalized-autoregressive-pretraining-for-language-understanding}
  {Xlnet: Generalized autoregressive pretraining for language understanding}.
\newblock In \emph{NIPS}, pages 5754--5764.

\bibitem[{Zhang et~al.(2020{\natexlab{a}})Zhang, Yuan, Liu, Fu, Zhuang, Wang,
  Chen, and Xiong}]{DBLP:journals/corr/abs-2009-02835}
Denghui Zhang, Zixuan Yuan, Yanchi Liu, Zuohui Fu, Fuzhen Zhuang, Pengyang
  Wang, Haifeng Chen, and Hui Xiong. 2020{\natexlab{a}}.
\newblock \href {http://arxiv.org/abs/2009.02835} {{E-BERT:} {A} phrase and
  product knowledge enhanced language model for e-commerce}.
\newblock \emph{CoRR}, abs/2009.02835.

\bibitem[{Zhang et~al.(2020{\natexlab{b}})Zhang, Jia, Yin, Dong, Gao, and
  Hua}]{DBLP:journals/corr/abs-2008-10813}
Ningyu Zhang, Qianghuai Jia, Kangping Yin, Liang Dong, Feng Gao, and Nengwei
  Hua. 2020{\natexlab{b}}.
\newblock \href {http://arxiv.org/abs/2008.10813} {Conceptualized
  representation learning for chinese biomedical text mining}.
\newblock \emph{CoRR}, abs/2008.10813.

\bibitem[{Zhang et~al.(2017)Zhang, Zhang, Wang, Cheng, Li, and
  Ding}]{zhang2017chinese}
Sheng Zhang, Xin Zhang, Hui Wang, Jiajun Cheng, Pei Li, and Zhaoyun Ding. 2017.
\newblock \href {https://www.mdpi.com/2076-3417/7/8/767} {Chinese medical
  question answer matching using end-to-end character-level multi-scale cnns}.
\newblock \emph{Applied Sciences}, 7(8):767.

\bibitem[{Zhang et~al.(2019)Zhang, Han, Liu, Jiang, Sun, and
  Liu}]{DBLP:conf/acl/ZhangHLJSL19}
Zhengyan Zhang, Xu~Han, Zhiyuan Liu, Xin Jiang, Maosong Sun, and Qun Liu. 2019.
\newblock \href {https://doi.org/10.18653/v1/p19-1139} {{ERNIE:} enhanced
  language representation with informative entities}.
\newblock In \emph{ACL}, pages 1441--1451.

\end{thebibliography}
\clearpage
\appendix
\section{Data Source}
\subsection{Pre-training Data}
\label{pre_training_data}
\subsubsection{Training Corpora}
The pre-training corpora is crawled from DXY BBS (Bulletin Board System) \footnote{\url{https://www.dxy.cn/bbs/newweb/pc/home}}, which is a very popular Chinese social network for doctors, medical institutions, life scientists, and medical practitioners. The BBS has more than 30 channels, which contains 18 forums and 130 fine-grained groups, covering most of the medical domains. 
For our pre-training purpose, we crawl texts from channels about clinical medicine, pharmacology, public health and consulting. For text pre-processing, we mainly follow the methods of \citep{DBLP:journals/corr/abs-2003-01355}. Additionally, (1) we remove all URLs, HTML tags, e-mail addresses, and all tokens except characters, digits, and punctuation (2) all documents shorter than 256 are discard, while documents longer than 512 are cut into shorter text segments.
\subsubsection{Knowledge Graph}
The DXY knowledge graph is construed by extracting structured text from DXY website\footnote{\url{https://portal.dxy.cn/}}, which includes information of diseases, drugs and hospitals edited by certified medical experts, thus the quality of the KG is guaranteed. The KG is mainly disease-centered, including totally 3,764,711 triples, 152.508 unique entities, and 44 types of relations.
The details of Symptom-In-Chinese from OpenKG is available \footnote{\url{http://openkg.cn/dataset/symptom-in-chinese}}.
We finally get 26 types of entities, 274,163 unique entities, 56 types of relations, and 4,390,726 triples after the fusion of the two KGs.
\subsection{Task Data}
\label{task_data}
We choose the four large-scale datasets in ChineseBlue tasks \citep{DBLP:journals/corr/abs-2008-10813} while others are ignored due to the limitation of datasets size, which are cMedQANER, cMedQQ, cMedQNLI and cMedQA. WebMedQA \citep{DBLP:journals/midm/HeFT19} is a real-world Chinese medical question answering dataset and CHIP-RE dataset are collected from online health consultancy websites.
Note that since both the WebMedQA and cMedQA datasets are very large while we have many baselines to be compared, we randomly sample the official training set, development set and test set respectively to form their corresponding smaller version for experiments.
DXY-NER and DXY-RE are datasets from real medical application scenarios provided by a prestigious Chinese medical company.
The DXY-NER contains 22 unique entity types and 56 relation types in the DXY-RE.
These two datasets are collected from the medical forum of DXY and books in the medical domain.
Annotators are selected from junior and senior students with clinical medical background.
In the process of quality control, the two datasets are annotated twice by different groups of annotators.
An expert with medical background performs quality check manually again when annotated results are inconsistent, whereas perform sampling quality check when results are consistent.
Table \ref{dataset_statistical_data} shows the datasets size of our experiments.

\begin{table*}[]
\normalsize
\centering
\begin{tabular}{m{3.4cm}<{\centering}  |m{1.5cm}<{\centering} m{1.5cm}<{\centering} m{1.5cm}<{\centering} m{2cm}<{\centering} m{1.5cm}<{\centering}}
\toprule
\multicolumn{6}{c}{The Dataset Size in Our Experiments} \\ \midrule
Dataset & Train & Dev & Test & Task & Metric            \\ \midrule
\makecell[c]{cMedQANER \\ \citep{DBLP:journals/corr/abs-2008-10813}}  & 1,673 & 175 & 215& NER& F1  \\ \midrule
\makecell[c]{cMedQQ \\ \citep{DBLP:journals/corr/abs-2008-10813}}  & 16,071 & 1,793 & 1,935 & QM & F1  \\ \midrule
\makecell[c]{cMedQNLI \\ \citep{DBLP:journals/corr/abs-2008-10813}}  & 80,950 & 9,065 & 9,969 & NLI & F1  \\ \midrule
\makecell[c]{cMedQA  \\ \citep{zhang2017chinese}}   & 186,771 & 46,600 & 46,600 & QA & Acc@1  \\ \midrule
\makecell[c]{WebMedQA \\ \citep{DBLP:journals/midm/HeFT19}}  & 252,850 & 31,605 & 31,655& QA & Acc@1  \\ \midrule
CHIP-RE $^*$  & 43,649 & - & 10,622 & RE & F1  \\ \midrule
DXY-NER  & 34,224 & 8,576 & 8,592& NER & F1  \\ \midrule
DXY-RE  & 141,696 & 35,456 & 35,794& RE & F1  \\ \bottomrule
\end{tabular}
\\ \footnotesize{ $^*$ CHIP-RE dataset is released in CHIP 2020. (\url{http://cips-chip.org.cn/2020/eval2})}\\
\caption{\label{dataset_statistical_data} The statistical data and metric of eight datasets used in our SMedBERT model.}
\end{table*}

\section{Model Settings and Training Details}
\label{model_para_traing_details}
\paragraph{Hyper-parameters.} $d_1$=768, $d_2$=200, $K$=10, $\mu$ =10, $\lambda_1$=2, $\lambda_2$=4.
\paragraph{Model Details.} We align the all mention-spans to the entity in KG by exact match for comparison purpose with ENIRE-THU \citep{DBLP:conf/acl/ZhangHLJSL19}. The negative sampling function is defined as $Q(e_m^{i})=\frac{t_{e_m^{i}}}{C_{e_m^{i}}}$, 
where $C_{e_m^{i}}$ is the sum of frequency of all mentions with the same type of $e_m^{i}$. The Mention-neighbor Hybrid Attention module is inserted after the tenth transformer encoder layer to compare with KnowBERT \citep{DBLP:conf/emnlp/PetersNLSJSS19}, while we perform the Mention-neighbor Context Modeling based on the output of BERT encoder.
We use all the base-version PLMs in the experiments. The size of SMedBERT is 474MB while 393MB of that are components of BERT, and the added 81MB is mostly of the KG embedding.
Results are presented in average with 5 random runs with different random seeds and the same hyper-parameters.

\paragraph{Training Procedure.} We strictly follow the originally pre-training process and parameter setting of other KEPLMs.
We only adapt their publicly available code from English to Chinese and use the knowledge embedding trained on our medical KG.
To have a fair comparison, the pre-training processing of SMedBERT is mostly set based on ENIRE-THU \citep{DBLP:conf/acl/ZhangHLJSL19} without layer-special learning rates in KnowBERT \citep{DBLP:conf/emnlp/PetersNLSJSS19}.
We only pre-train SMedBERT on the collected medical data for 1 epoch.
In pre-training process, the learning rate is set to $5e^{-5}$ and batch size is 512 with the max sequence length is 512.
For fine-tuning, we find the following ranges of possible values work well, i.e., batch size is \{8,16\}, learning rate (AdamW) is \{$2e^{-5}$, $4e^{-5}$, $6e^{-5}$\} and the number of epochs is \{2,3,4\}.
Pre-training SMedBERT takes about 36 hours per epoch on 2 NVIDIA GeForce RTX 3090 GPUs.

\section{Data and Embedding of Unsupervised Semantic Similarity}
\label{unsupervised_semantic_similarity}
Since the KGs used in this paper is a directed graph, we first transform the directed "等价关系" (equivalence relations) pairs to undirected pairs and discard the duplicated pairs. For each positive pairs, we use head and tail as query respectively and sample the negative candidates based on the other. Specifically, we randomly select 19 negative entities with the same type and has a Jaro-Winkle similarity \citep{winkler1990string} bigger 0.6 with the ground-truth entity. We select from all samples in \textbf{Dataset-1} with positive pairs that the neighbours sets of head and tail entity have Jaccard Index \citep{2010THE} no less than 0.75 and at least 3 common element to construct the \textbf{Dataset-2}.
For \textbf{Dataset-3}, we count the frequency of all entity mentions in pre-training corpora, and treat mentions with frequency no more than 200 as low-frequency mentions.
\paragraph{Classic Word Representation Embedding:} We train the character-level and word-level embedding using SGNS \citep{DBLP:journals/corr/abs-1301-3781} and GLOVE \citep{DBLP:conf/emnlp/PenningtonSM14} model respectively on our medical corpora with open-source toolkits\footnote{SGNS: \url{https://github.com/JuGyang/word2vec-SGNS}. \\ Glove: \url{https://github.com/stanfordnlp/GloVe}}. 
We average the character embedding for all tokens in the mention to get the character-level representation.
However, since some mentions are very rare in the corpora for word-level representation, we use the character-level representation as their word-level representation.
\paragraph{BERT-like Representation Embedding:} We extract the token hidden features of the last layer and average the representations of the input tokens except [CLS] and [SEP] tag, to get a vector for each entity.
\paragraph{Similarity Measure:} We try using the inverse of L2-distance and cosine similarity as measurement, and we find that cosine similarity always perform better. Hence, we report all experiment results under the cosine similarity metric.

\end{CJK}
\end{document}